\documentclass[11pt]{article}

\usepackage[preprint]{acl}

\usepackage{booktabs}
\usepackage{times}
\usepackage{latexsym}

\usepackage[T1]{fontenc}

\usepackage[utf8]{inputenc}

\usepackage{microtype}
\usepackage{makecell}
\usepackage{inconsolata}

\usepackage{graphicx}
\usepackage{arydshln}
\usepackage{paralist}

\usepackage{fontawesome5}
\usepackage{lipsum}

\usepackage[most]{tcolorbox}
\usepackage{listings}
\usepackage{xcolor}
\usepackage{amsfonts}
\definecolor{headergray}{RGB}{245, 245, 245}
\definecolor{codebg}{RGB}{250, 250, 250}
\definecolor{interpreterbg}{RGB}{240, 240, 255} 
\definecolor{AcademicGreen}{HTML}{004225}
\definecolor{AcademicRed}{HTML}{A51C30}
\title{DuDi: Dual-Signal Distillation with Cross-Lingual Verbalizer}

\author{
 \textbf{Patomporn Payoungkhamdee\textsuperscript{1*$\dagger$}},
 \textbf{Tinnakit Udsa\textsuperscript{1*}},
 \textbf{Jian Gang Ngui\textsuperscript{2}},
 \\
 \textbf{Sarana Nutanong\textsuperscript{1}},
 \textbf{Alham Fikri Aji\textsuperscript{3}},
 \textbf{Peerat Limkonchotiwat\textsuperscript{2}}
\\
 \textsuperscript{1}School of Information Science and Technology, VISTEC
 \textsuperscript{2}AI Singapore
 \textsuperscript{3}MBZUAI
 \\
 \texttt{\{patomporn.p\_s21,tinnakit.u\_s24\}@vistec.ac.th},
 \texttt{peerat@aisingapore.org}
 \\
 \href{https://github.com/aisingapore/DuDi}{\faGithub\ GitHub}
\quad
\href{https://huggingface.co/collections/aisingapore/sea-lion-research}{\includegraphics[height=1em]{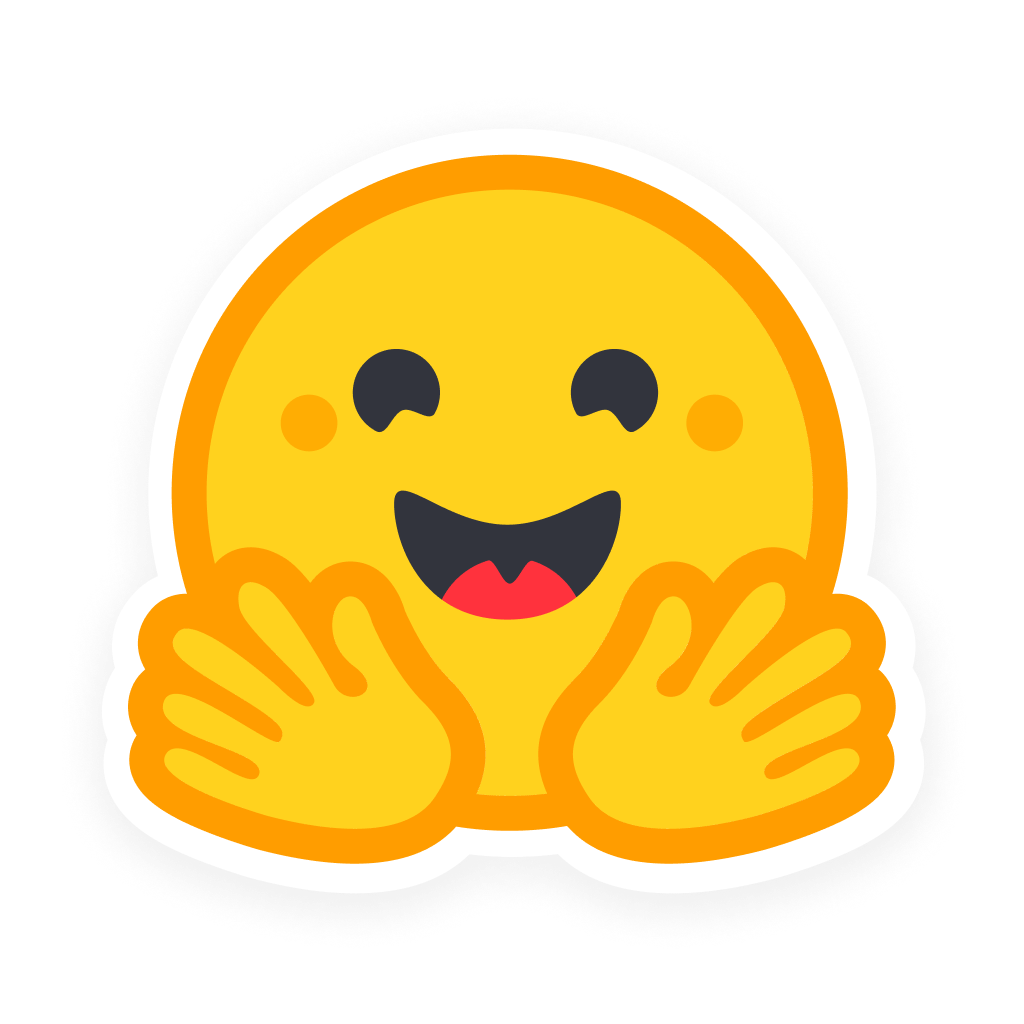}\ Hugging Face}
}

\begin{document}
\maketitle
\begingroup
\renewcommand\thefootnote{*}
\footnotetext{Equal contribution}
\endgroup
\begingroup
\renewcommand\thefootnote{$\dagger$}
\footnotetext{Work was conducted while Patomporn Payoungkhamdee was a visiting scholar at AI Singapore}
\endgroup
\begin{abstract}

Small language models (SLMs) are efficient and scalable, but their multilingual capabilities degrade severely at sub-billion scales, especially for Southeast Asian (SEA) languages.
We introduce DuDi, a dual-signal multilingual distillation framework that combines an online sequence-level signal with off-policy and on-policy token-level signals.
DuDi further uses a cross-lingual verbalizer to refine teacher feedback and improve teacher-student transferability in multilingual settings.
Experiments on SEA-HELM across multiple model families, scales, and teacher--student settings show that DuDi consistently outperforms competitive distillation baselines.
Ablations and analyses confirm that sequence-level optimization, token-level supervision, and cross-lingual verbalization provide complementary and transferable learning signals for multilingual SLMs.

\end{abstract}

\section{Introduction}

Small language models (SLMs) have recently attracted growing attention due to their efficiency and scalability~\citep{minicpm,survey_slm2,survey_slm3,survey_slm4}.
For instance, Qwen2.5-1.5B and -0.5B achieve about 1.5$\times$ and 2.2$\times$ higher inference throughput than Qwen2.5-7B, while reducing memory use by 48\% and 81\%, respectively.\footnote{\tiny Measured with BF16 precision and input length 6144, based on benchmarks from \url{https://qwen.readthedocs.io/en/v2.5/benchmark/speed_benchmark.html}}
These gains reduce deployment costs and enable more efficient large-scale serving.
Compared with larger models, SLMs offer practical advantages in scalability, computation, and memory usage~\citep{minicpm}.
These properties make them suitable for resource-constrained and edge-device deployment~\citep{mobilellm}, while supporting real-world applications at scale~\citep{slimlm,c2kd-slm-rec}.

\begin{figure}[tb]
\centering
\includegraphics[width=\columnwidth]{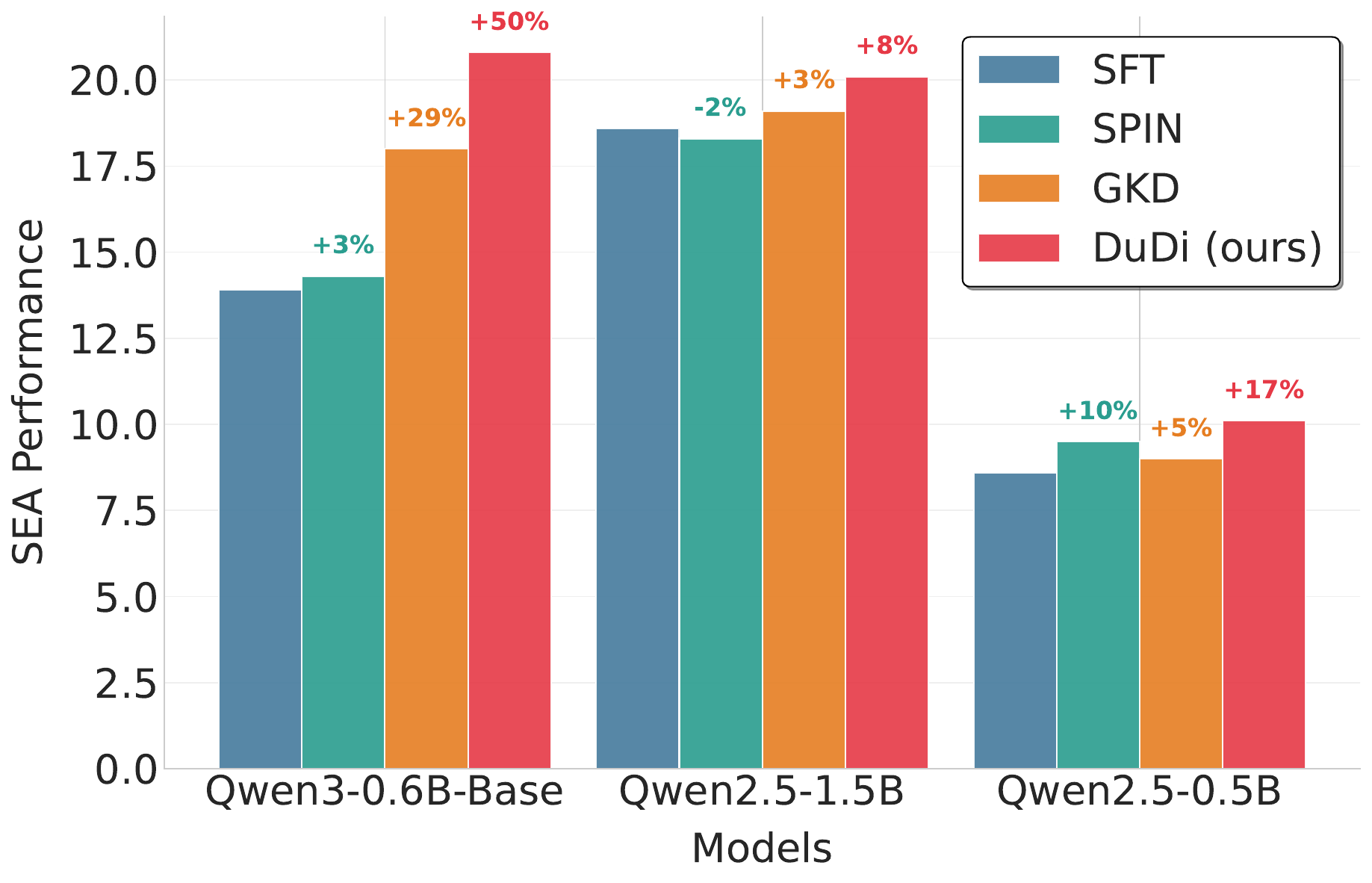}
    \caption{
    Comparison of SEA performance across different model scales and training frameworks, evaluated using the SEA-HELM benchmark ($\uparrow$). Details of each model are provided in Section~\ref{sec:setup}.
    }
    \vspace{-5mm}
    \label{fig:perf-scaling-trend}
\end{figure}

However, multilingual capabilities in SLMs remain limited~\cite{survey-multilingual-llm,mmlu-prox}, especially for Southeast Asian (SEA) languages, a highly diverse region with hundreds of millions of speakers.
As shown in Figure~\ref{fig:perf-scaling-trend}, SEA performance drops substantially as model size falls below the billion-scale regime.
In particular, the Supervised Fine-Tuning (SFT) variant of Qwen2.5-0.5B drops sharply relative to Qwen2.5-1.5B, while the newer Qwen3-0.6B still shows limited SEA performance under standard SFT.
These results suggest that scaling down weakens multilingual understanding, motivating training strategies tailored for SLMs.

A common approach to improve SLMs' performance is knowledge distillation (KD), which transfers knowledge from a larger teacher to a smaller student~\cite{hintonkd,seqkd,gkd,gu2024minillm,DistiLLM,DistiLLM-2}.
Despite recent advances in KD, multilingual distillation remains largely limited to task-specific or data-centric settings~\cite{d-cot-nr,SDRRL-multi-self-distill-resource}, leaving general-purpose multilingual distillation for SLMs underexplored.
%
%
%
%
This gap is especially pronounced for SEA languages due to their linguistic diversity and limited high-quality training data, motivating multilingual KD strategies designed for SLMs.

To address this limitation, we propose \textbf{Du}al-Signal \textbf{Di}stillation with Cross-Lingual Verbalizer (DuDi), a general-purpose framework designed for multilingual distillation in SLMs.
DuDi builds a unified KD framework around three components: \emph{Sequence Signal}, \emph{Token Signal}, and \emph{Cross-Lingual Verbalizer}.
For the sequence signal, DuDi introduces an online sequence-level objective that guides the student policy toward the ground-truth direction.
For the token signal, DuDi uses both off-policy and on-policy supervision: off-policy signals come from the training corpus, while on-policy signals come from student-generated responses.
To facilitate knowledge transferability, DuDi uses a cross-lingual verbalizer to refine teacher logits during on-policy distillation, aligning student responses with the ground-truth demonstrations.
This design enables us to better facilitate student learning in a multilingual environment.

To evaluate DuDi, we compare it with competitive methods under the SEA training and evaluation framework, using SEA-Instruct\footnote{\tiny \url{https://huggingface.co/datasets/aisingapore/SEA-Instruct-2602}} and SEA-HELM~\cite{sea-helm}.
The experimental results demonstrate that DuDi achieves the strongest overall performance under the Qwen2.5-0.5B setting, with gains across most SEA languages.
This trend generalizes across scales and architectures, demonstrating scalability and robustness.
Ablations show consistent degradation when any DuDi component is removed, highlighting the need to jointly optimize sequence-level objective, dual-policy token signals, and the cross-lingual verbalizer.
Finally, analysis of the DuDi verbalizer demonstrates that it provides richer learning signals for teacher-student distillation.
%

In conclusion, our contributions are as follows:
\begin{compactitem}
    \item We propose \textbf{DuDi}, a multilingual knowledge distillation framework that integrates sequence-level signals and token-level, improving SEA performance in small LMs.
    \item We introduce a \textbf{cross-lingual verbalizer} that better facilitates on-policy distillation.
    \item We conduct \textbf{ablations and analyses} to assess each component, showing the effectiveness of dual-signal and verbalizer designs.
\end{compactitem}

\section{Background}\label{subsec:background}

\begin{table*}[h!]
\centering
\small
\setlength{\tabcolsep}{6pt}
\renewcommand{\arraystretch}{1.2}
\resizebox{\textwidth}{!}{
\begin{tabular}{l l c c c c}
\hline
\textbf{Method} & \textbf{Teacher} & \textbf{Off-Policy Token-Signal} & \textbf{On-Policy Token-Signal} & \textbf{Sequence-Signal} & \textbf{Verbalizer} \\
\hline
SFT    & $\times$ & $\checkmark$ & $\times$ & $\times$ & $\times$ \\
DFT~\cite{dft}   & $\times$ & $\checkmark$ & $\times$ & $\times$ & $\times$ \\
SPIN~\cite{spinwork}  & $\times$ & $\times$ & $\times$ & $\checkmark$  & $\times$ \\
SDFT~\cite{sdft}  & Self & $\times$   & $\checkmark$ & $\times$ & English \\
SeqKD~\cite{seqkd} & Larger & $\checkmark$ & $\times$ & $\times$ & $\times$ \\
GKD~\cite{gkd}   & Larger & $\checkmark$ & $\checkmark$ & $\times$ & $\times$ \\
DuDi (Ours)  & Larger & $\checkmark$ & $\checkmark$ & $\checkmark$ & Cross-lingual \\
\hline
\end{tabular}
}
\vspace{-3mm}
\caption{Comparison of training paradigms across different frameworks.}
\vspace{-6mm}
\label{tab:method-comparison}
\end{table*}

The multilingual training corpus consists of an input $x$, a ground-truth $y$, and a language $l$. Formally, this dataset is defined as $\mathcal{D} = \{(x_i, y_i, l_i)\}_{i=1}^N$, where $N$ denotes the total number of training samples.
To learn from the data, a fine-tuning methodology is employed to optimize the policy $\pi_\theta$. 
This process involves minimizing an objective function, denoted as $\mathcal{L}(x_i, y_i, l_i; \pi_\theta)$, which serves as a metric for the difference between the model's stochastic predictions and the ground-truths.
Existing methods structure this objective differently to address distinct learning dynamics.

\noindent \textbf{Off-Policy Fine-Tuning.} 
This method represents a straightforward approach, typically grounded in a cross-entropy objective. Given a model policy $\pi_\theta$, the loss function is formulated as
\begin{equation}
    \mathcal{L}_{\text{Off-FT}} = \mathbb{E}_{(x,y)\sim\mathcal{D}}\left[- w \log \pi_\theta(y| x)\right],
\end{equation}
where $w$ serves as a weighting coefficient to modulate the training signal. 
In standard Supervised Fine-Tuning (SFT), $w = 1$, treating all tokens with equal importance. 
In Dynamic Fine-Tuning (DFT) \cite{dft}, $w$ is defined as $\text{sg}(\pi_\theta(y|x))$, where $\text{sg}(\cdot)$ denotes the stop-gradient operator. This token-weighting is designed to stabilize gradient magnitudes and improve generalization during the fine-tuning process.

\noindent \textbf{Iterative Fine-Tuning.}
To address the limitation of static training data, prior work has explored iterative self-play for policy refinement \cite{td-gammon,alphago}. 
\citet{spinwork} proposed Self-Play Fine-Tuning (SPIN), a bootstrapping framework that improves the model by distinguishing ground-truths and self-generated responses sampled from an SFT-initialized reference policy $y' \sim \pi_{\theta_\text{Ref}}(x)$.
This approach optimizes the policy by maximizing an Integral Probability Metric against a previous iteration of the self. The iterative fine-tuning objective is defined as
\begin{equation}
\resizebox{\columnwidth}{!}{$
    \mathcal{L}_\text{IFT} = \mathbb{E}_{(x,y)\sim\mathcal{D},y'}\left[\ell\left(\lambda\log\frac{\pi_{\theta_\text{Ref}}(y|x)\pi_{\theta}(y'|x)}{\pi_{\theta}(y|x)\pi_{\theta_\text{Ref}}(y'|x)}\right)\right],   
$}
\label{eq:spin}
\end{equation}
where $\ell(t) = \log(1 + \exp(-t))$ is the logistic loss and $\lambda > 0$ is the regularization parameter. By contrasting the log-likelihood ratios of target responses against its own generations, the model increasingly aligns its policy with the ground-truth distribution through successive iterations.


\noindent \textbf{Self-Distillation.}
To mitigate the generalization of off-policy fine-tuning, several studies adopt a self-distillation paradigm \cite{sdft-acl,SDRRL-multi-self-distill-resource,sdpo}.
Specifically, \citet{sdft} introduced Self-Distillation Fine-Tuning (SDFT), which transforms off-policy signals into an on-policy paradigm by employing an exponential moving average student as the teacher $\pi_{\theta_\text{T}}\sim\text{EMA}(\pi_\theta)$.
A central component of SDFT is an English verbalizer function, $z\sim v_\text{en}(x, y)$, which converts an input and ground-truth pair into a structured demonstration prompt for the teacher model (illustrated in Figure~\ref{fig:verbalizer_template}). This verbalized demonstration is subsequently used to guide the student-generated response, $\tilde{y}\sim\pi_\theta(x)$.
The optimization objective minimizes the divergence between the teacher, conditioned on the verbalized demonstration, and the student policy:
\begin{equation}
    \mathcal{L}_\text{SD} = \mathbb{E}_{x\sim{D},\tilde{y},z} \left[D\left(\pi_{\theta_\text{T}}(\tilde{y}|z)||\pi_{\theta}(\tilde{y}|x) \right)\right].
\end{equation}
By leveraging this temporary ensemble, SDFT regularizes the optimization path and improves generalization through token-level guidance.

\noindent \textbf{Teacher Distillation.} 
The teacher knowledge distillation \cite{hintonkd,ImitKD,DistiLLM,DistiLLM-2} paradigm leverages signals from a stronger teacher model ($\pi_{\theta_{\text{T}}}$) to guide a student model ($\pi_\theta$), typically a parameter-efficient counterpart.
This approach abstractly optimizes two objectives that could utilize static ground-truths with stochastic explorations:
\begin{equation}
\resizebox{\columnwidth}{!}{$
\begin{aligned}
    \mathcal{L}_\text{TD} = &(1-\lambda) \mathbb{E}_{(x,y)\sim\mathcal{D}}\left[D(\pi_{\theta_\text{T}}(y|x)||\pi_{\theta}(y|x))\right] \\
    &+\lambda \mathbb{E}_{x\sim\mathcal{D},\tilde{y}\sim\pi_\theta(x)}\left[D\left(\pi_{\theta_\text{T}}(\tilde{y}|x)||\pi_{\theta}(\tilde{y}|x) \right)\right],
\end{aligned}
$}
\label{eq:kd}
\end{equation}
where $D$ denotes a divergence function, and $\lambda~\in~[0,1]$ balances the distillation signals from ground-truths and newly-generated responses.
Here, $\tilde{y}$ is sampled from either the teacher or student policy. In off-policy KD, SeqKD \cite{seqkd} trains the student on teacher-generated sequences, i.e., $\tilde{y} \sim \pi_{\theta_{\mathrm{T}}}(x)$.
However, this approach often suffers from training-inference mismatches when the output sequences generated by the student at inference time deviate significantly from those encountered during training.
To address this mismatch, Generalized Knowledge Distillation, GKD \cite{gkd} introduces an on-policy mechanism. In the GKD framework, the student generates its own responses $\tilde{y} \sim \pi_\theta(x)$, while the teacher provides token-level signals for guiding the student output logits to correct their self-generated mistakes.
%


As summarized in Table~\ref{tab:method-comparison}, although these approaches have advanced fine-tuning methodologies, their multilingual extension remains insufficiently explored.
Additionally, existing methods typically treat sequence-level and token-level supervision independently, limiting the complementarity of both learning signals within a unified framework.

\section{DuDi}
\label{sec:DuDi}

\begin{figure}[h!]
\centering
\includegraphics[width=\columnwidth]{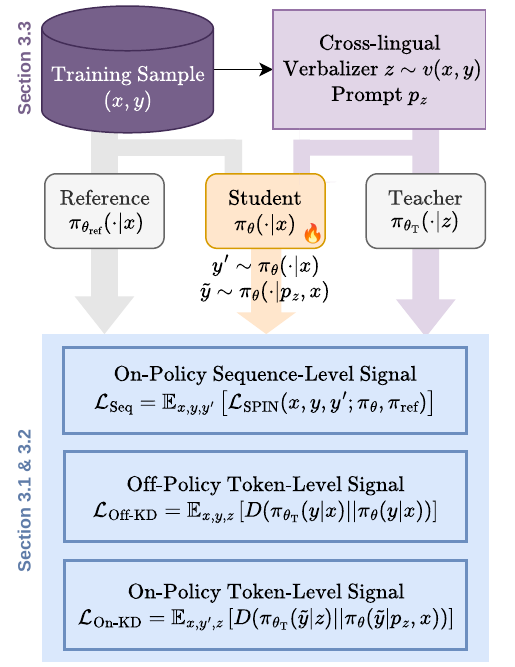}
    \caption{
    Overview of the DuDi framework, which integrates the sequence-level objective, token-level signals of off-policy, and on-policy knowledge distillation with teacher-guided cross-lingual verbalization.
    }
    \vspace{-5mm}
    \label{fig:DuDi}
\end{figure}

%
As illustrated in Figure~\ref{fig:DuDi}, DuDi consists of three core components: a sequence-level signal (Section~\ref{sec:seq-signal}), a token-level signal (Section~\ref{sec:tok-signal}), and a cross-lingual verbalizer (Section~\ref{sec:cross-lingual-verb}).
The framework enables SLMs to jointly leverage sequence-level and token-level supervision, while the cross-lingual verbalizer improves teacher-student knowledge transferability in multilingual settings.
%

\subsection{Sequence Signal}
\label{sec:seq-signal}

 To improve the efficiency of SLMs, we integrate a sequence-level objective inspired by the SPIN framework to ensure the student policy converges toward the ground-truth demonstration. This objective requires the model to differentiate its current policy from the ground-truth by maximizing the relative likelihood of ground-truths. Distinct from static iterations of SPIN, we sample responses $y'\sim\pi_\theta(x)$ in \textit{real-time} to provide dynamic on-policy feedback with the objective in Equation~\ref{eq:spin}.


\subsection{Token Signal}
\label{sec:tok-signal}

Complementing token-level supervision, DuDi leverages both off-policy and on-policy objectives to enhance performance through knowledge distillation.

\noindent \textbf{Off-Policy  KD.}
While on-policy distillation mitigates teacher-student mismatches~\citep{gkd,gu2024minillm,DistiLLM,DistiLLM-2}, relying solely on student-generated responses may reduce exposure to the ground-truth data distribution, potentially causing the learned policy to drift away from it. To address this problem, we adopt an off-policy distillation signal that leverages teacher-provided logits as guidance. This mechanism ensures that the supervisory signal remains strictly grounded in the ground-truth distribution. The corresponding objective can be represented as
\begin{equation}
\begin{split}
    \mathcal{L}_\text{Off-KD} &= D\left(\pi_{\theta_\text{T}}(y|x)||\pi_{\theta}(y|x) \right).
\end{split}
\label{eq:off_policy_tok}
\end{equation}

\noindent \textbf{On-policy KD.}
To enable teacher-guided refinement of the student-generated responses, we further adopt an on-policy knowledge distillation objective, where the student generates a response $\tilde{y}\sim\pi_\theta(x)$.
The on-policy distillation objective is defined as
\begin{equation}
\begin{split}
    \mathcal{L}_\text{On-KD} &= D\left(\pi_{\theta_\text{T}}(\tilde{y}|x)||\pi_{\theta}(\tilde{y}|x) \right).
\end{split}
\label{eq:on_policy_tok}
\end{equation}

%

\subsection{Cross-Lingual Verbalizer}
\label{sec:cross-lingual-verb}

\begin{figure}[h!]
    \vspace{-1mm}
\centering
\includegraphics[width=\columnwidth]{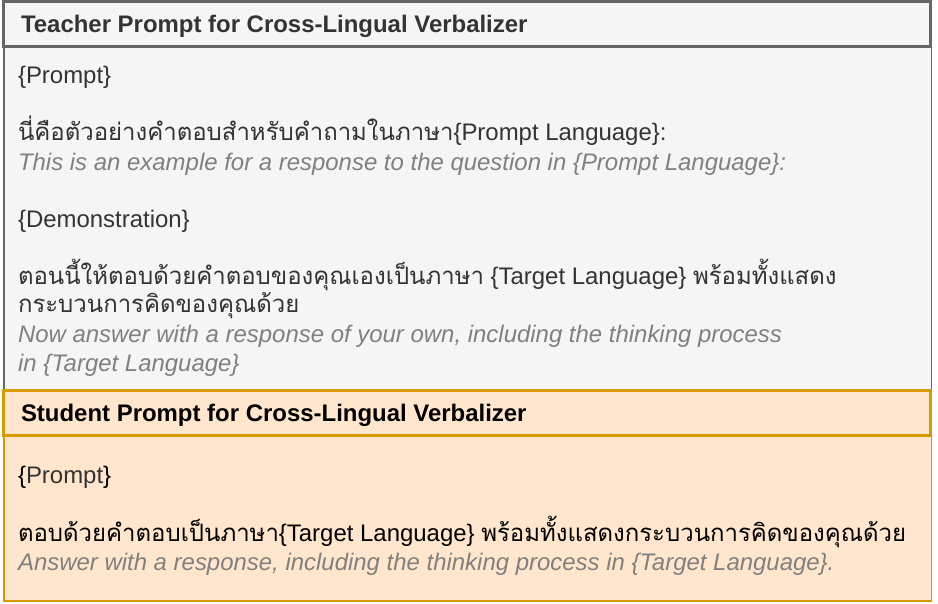}
    \vspace{-6mm}
    \caption{Illustration of the cross-lingual verbalizer template, showing the teacher ($z$) and its corresponding student prompt ($p_z$) from a Thai training instance.}
    \vspace{-7mm}
    \label{fig:cross_linugal_verbalizer_template}   
\end{figure}

\begin{figure}[h!]
\centering\includegraphics[width=\columnwidth]{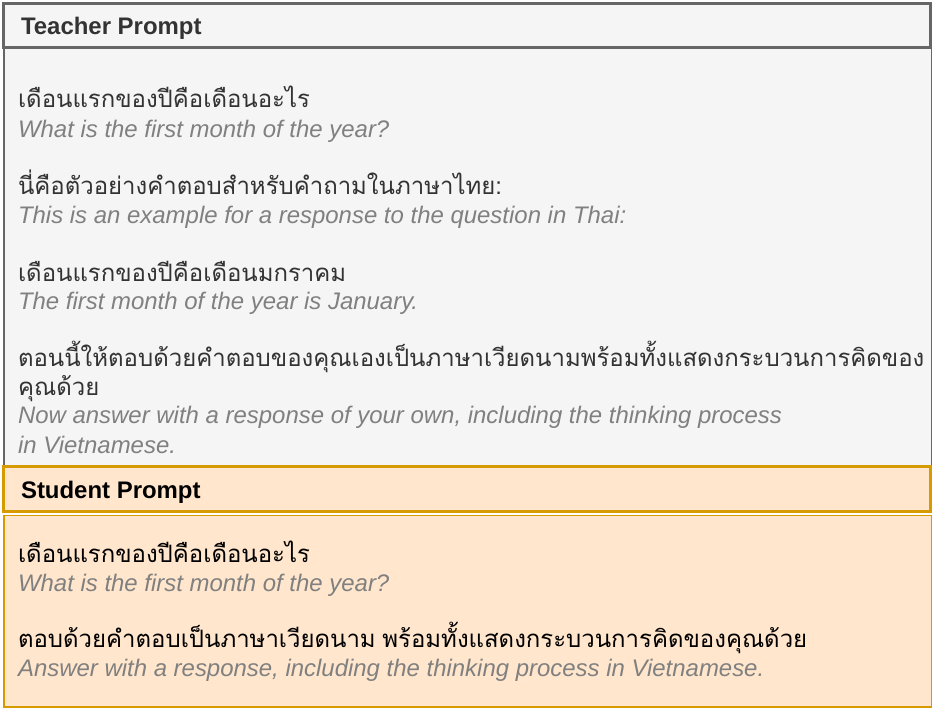}
    \caption{Example of cross-lingual verbalized teacher and student prompts. In this example, the original sample is in Thai, where the prompt is expressed in Thai, while the target response is generated in Vietnamese. For this sample, the target response language is uniformly sampled from seven languages: English, Indonesian, Vietnamese, Tamil, Tagalog, Malay, and Burmese, excluding the sample’s original language (Thai).}
    \label{fig:verbalizer_example}    
    \vspace{-4mm}
\end{figure}

Optimizing on-policy learning with token-level supervision in multilingual settings requires a carefully designed framework for effective teacher-student knowledge transfer.
To further improve the distillation process, inspired by English verbalizer~\citep{sdft}, we introduce a cross-lingual verbalizer.
This component converts an input prompt $x$, a ground-truth $y$ as demonstration, source language $l$, and target language $l_z$ into a verbalized prompt $z = v(x, y, l, l_z)$ for the teacher, along with a corresponding prompt $p_z$ for the student. 
The verbalizer prompt is the same language $l$ as the sample native language. The prompt $p_z$ instructs the student model to generate responses in the target language $l_z$, where $l_z$ sampled uniformly from the set of training languages excluding $l$, with English included as an additional language.
An example of the cross-lingual verbalizer prompt template is shown in Figure~\ref{fig:cross_linugal_verbalizer_template}.
Furthermore, Figure~\ref{fig:verbalizer_example} illustrates a Thai sample in which verbalized for generating responses in Vietnamese.

%

%

Consequently, with a cross-lingual verbalizer, the on-policy token-level distillation objective in Equation~\ref{eq:on_policy_tok} could be modified with a cross-lingual verbalizer $z$ for the teacher together with its associated prompt template $p_z$ for the student. Based on this verbalized input, the student generates an additional response $\tilde{y}\sim\pi_\theta(p_z,x)$.
%
%
The on-policy distillation objective with the cross-lingual verbalizer on the teacher is formally expressed as:
\vspace{-1mm}
\begin{equation}
\begin{split}
    \mathcal{L}_\text{On-KD} &= D\left(\pi_{\theta_\text{T}}(\tilde{y}|z)||\pi_{\theta}(\tilde{y}|p_z,x) \right).
\end{split}
\label{eq:on_policy_tok_verbal}
\end{equation}
\vspace{-5mm}

The use of the cross-lingual verbalizer enables knowledge transfer across languages, thereby improving downstream performance in multilingual settings. The cross-lingual verbalizer configuration in DuDi is detailed in Section~\ref{ssec:setup}.

\subsection{Unify Training Objective} 
The DuDi framework is optimized by integrating on-policy sequence-level alignment (Equation~\ref{eq:spin}), off-policy token-level distillation (Equation~\ref{eq:off_policy_tok}), and on-policy token-level guidance via the multilingual verbalizer (Equation~\ref{eq:on_policy_tok_verbal}). Formally, the learning objective is abstractly represented as:
\vspace{-1mm}
\begin{equation}
\resizebox{\columnwidth}{!}{$
\begin{aligned}
    &\mathcal{L}_\text{DuDi} = \alpha\mathbb{E}_{(x,y)\sim\mathcal{D},y'\sim\pi_\theta(x)}\left[\mathcal{L}_\text{SPIN}(x,y,y';\pi_\theta,\pi_{\theta_\text{Ref}})\right] \\
    &+ 
    (1-\lambda)\mathbb{E}_{(x,y)\sim\mathcal{D}}\left[\mathcal{L}_\text{Off-KD}(x,y;\pi_\theta,\pi_{\theta_\text{T}})\right]  \\ 
    &+ \lambda\mathbb{E}_{x\sim\mathcal{D},\tilde{y}\sim
    \pi_\theta(p_z,x),z\sim v(x,y)}\left[\mathcal{L}_\text{On-KD}(x,\tilde{y},z;\pi_\theta,\pi_{\theta_\text{T}})\right],
\end{aligned}
$}
\end{equation}
\vspace{-2mm}

where $\alpha$ denotes the weighting coefficient for sequence-level student policy optimization and $\lambda$ denotes the weighting coefficient to balance token-level off-policy and on-policy distillation loss.
This joint objective allows the student model to leverage both the sequence-level optimization towards the ground-truth data distribution and fine-grained token-level supervision from the teacher.
%

\subsection{Differentiation from Previous Work}

DuDi jointly leverages sequence-level and token-level signals, with two key distinctions from prior approaches.
We integrate a self-play mechanism \cite{spinwork} as a sequence-level signal, transitioning from offline to online generation to reflect the student’s evolving policy.
Furthermore, we extend the on-policy distillation objective \cite{gkd,sdft} by introducing cross-lingual prompting for student rollouts, paired with a teacher equipped with a cross-lingual verbalizer, thereby improving knowledge transferability between teacher and student.

Compared to prior methods summarized in Table~\ref{tab:method-comparison}, this design enables the integration of dual-signal supervision with a tailored verbalizer, achieving state-of-the-art performance in multilingual settings (Section~\ref{sec:main_result}).
Ablation results in Section~\ref{sec:ablation-components} further show that these components are not independently effective but require joint integration in multilingual scenarios.
Moreover, Section~\ref{sec:analysis} demonstrates that the cross-lingual verbalizer improves on-policy teacher-student knowledge transferability, consistent with our hypothesis.

\section{Experimental Setup}
\label{sec:setup}

\begin{table*}[h!]
  \centering
  \scalebox{0.7}{
  \begin{tabular}{l|lllllll|l}
    \hline
    \textbf{Method} & \textbf{Indonesian} & \textbf{Vietnamese} & \textbf{Thai} & \textbf{Tamil} & \textbf{Tagalog} & \textbf{Malay} & \textbf{Burmese} & \textbf{Average}\\
    \hline
    Qwen2.5-3B-Instruct (Teacher) & 42.0 & 38.7 & 32.3 & 9.8 & 24.6 & 40.0 & 6.5 & 27.7 \\
    \hline
    \multicolumn{1}{l|}{Qwen2.5-0.5B (Student)} & \multicolumn{7}{c|}{} & \multicolumn{1}{c}{}\\
    SFT & 10.6  & 11.8  &  10.8 & 4.9  & 5.6  & 12.0  & 4.4  & 8.6  \\
    DFT & 9.3  &  11.6 & 10.2  & \textbf{10.0}  & 6.1  & 12.4  & \textbf{8.3}  & \underline{9.7}  \\
    \hdashline    
    SPIN & \textbf{12.6}  & 11.4  & 10.0  & 5.3  & \textbf{8.0}  & \underline{14.7}  & \underline{4.8}  & 9.5  \\ 
    \hdashline    
    SDFT & 4.2 & 4.6  & 4.7  & 2.3  & 3.3  & 5.1  & 1.3  & 3.6  \\
    SeqKD & 7.1 & 9.1  & 7.7  & 4.2  & 3.4  & 8.1  & 2.9  &  6.1 \\
    GKD & 11.7  & \underline{13.2}  & \underline{10.9}  & 4.9  & 4.8  & 13.6  & 3.7  & 9.0  \\
    \hdashline
    DuDi (ours) & \underline{11.7}$^\diamond$  & \textbf{14.4}$^\diamond$  & \textbf{12.8}$^\diamond$  & \underline{6.5}  & \underline{6.1}  & \textbf{14.8}$^\diamond$  & 4.6  & \textbf{10.1}$^\diamond$  \\
    \hline
  \end{tabular}
  }
  \vspace{-3mm}
  \caption{
  Downstream evaluation of methods across seven Southeast Asian languages. "$\diamond$" denotes a statistically significant improvement in each language, where DuDi outperforms DFT, the second-best overall performance.
  }
  \vspace{-3mm}
  \label{tab:main-result}
\end{table*}

\begin{table*}[h!]
  \centering
  \scalebox{0.7}{
  \begin{tabular}{l|lllllll|l}
    \hline
    \textbf{Method} & \textbf{Indonesian} & \textbf{Vietnamese} & \textbf{Thai} & \textbf{Tamil} & \textbf{Tagalog} & \textbf{Malay} & \textbf{Burmese} & \textbf{Average}\\
    \hline
    Qwen2.5-3B-Instruct (Teacher) & 42.0 & 38.7 & 32.3 & 9.8 & 24.6 & 40.0 & 6.5 & 27.7 \\
    \hdashline    
    \multicolumn{1}{l|}{Qwen2.5-1.5B (Student)} & \multicolumn{7}{c|}{} & \multicolumn{1}{c}{}\\
    SFT & 21.4 & 26.0  & \textbf{23.8}  & \underline{10.0}  & 16.6  & 26.1  & \textbf{6.2}  &  18.6 \\
    DFT & 22.1  & 17.8  & 16.6  & 9.6  & 13.6  & 22.8  & \textbf{6.2}  &  15.6 \\    
    SPIN & 21.0  & 25.3  & \underline{20.2}  & \textbf{10.1} & \textbf{19.2}  & 26.7  & \underline{5.3}  &  18.3 \\
    GKD & \textbf{28.6} & \underline{28.8}  & 19.9  & 7.3  & 15.4  & \underline{29.1}  & 4.7  &  \underline{19.1} \\
    DuDi & \underline{27.9}  & \textbf{30.3}$^\diamond$  & 19.8  & 8.5$^\diamond$  & \underline{19.0}$^\diamond$  & \textbf{30.7}$^\diamond$  & 4.8  &  \textbf{20.1}$^\diamond$ \\
    \hline    
    Qwen3-4B (Teacher) & 54.1 & 52.7  & 50.7  & 43.1  & 45.9  & 53.0  & 20.0  & 45.6 \\
    \hdashline
    \multicolumn{1}{l|}{Qwen3-0.6B-Base (Student)} & \multicolumn{7}{c|}{} & \multicolumn{1}{c}{}\\
    SFT & 15.6 & 19.5  & 17.3  & 10.1  & 10.7  & 17.9  & 6.4  & 13.9  \\
    DFT &  18.4 & 18.1  & 17.3  & \textbf{14.0}  & 14.0  & 20.1  & 7.1  & 15.6  \\    
    SPIN & 14.4 & 20.5  & 17.7  & 10.4  & 14.0  & 16.1  & \underline{6.8}  & 14.3  \\
    GKD & \underline{20.6} & \underline{27.0}  & \underline{21.9}  & 10.9  & \underline{15.1}  & \underline{23.9}  & 6.5  &  \underline{18.0} \\
    DuDi & \textbf{24.2}$^\diamond$  & \textbf{30.4}$^\diamond$  & \textbf{23.4}$^\diamond$  & \underline{13.2}$^\diamond$  & \textbf{17.6}$^\diamond$  & \textbf{28.1}$^\diamond$  & \textbf{8.4}$^\diamond$  & \textbf{20.8}$^\diamond$   \\
    \hline
    Llama3.2-3B-Instruct (Teacher) & 32.4 & 30.3 & 38.2 & 16.7 & 28.3 & 42.8 & 5.0 & 27.7 \\
    \hdashline
    Llama3.2-1B (Student) &   &   &   &   &   &   &   &   \\
    SFT & 4.7 & 5.3 & 3.6 & 3.9 & 3.3 & 6.5 & \textbf{5.7} & 4.7 \\
    
    DFT & 0.4 & 0.4 & 0.9 & 0.4 & 0.3 & 1.1 & 0.4 & 0.6 \\
    
    SPIN & 2.9 & 3.6 & 3.8 & 3.1 & 2.9 & 7.4 & 2.4 & 3.7 \\
    
    GKD & \underline{11.8} & \textbf{17.7} & \textbf{11.4} & \underline{5.6} & \underline{7.3} & \underline{16.3} & 4.2 & \underline{10.6} \\
    
    DuDi & \textbf{14.4}$^\diamond$ & \underline{16.5} & \underline{10.4} & \textbf{6.6}$^\diamond$ & \textbf{8.1} & \textbf{18.5}$^\diamond$ & \underline{5.2}$^\diamond$ & \textbf{11.4}$^\diamond$ \\
    \hline        
  \end{tabular}
  }
  \vspace{-3mm}
  \caption{
  Results across different teacher-student model configurations. "$\diamond$" denotes a statistically significant improvement in each language, where DuDi outperforms GKD, the second-best overall performing framework.
  }
  \vspace{-5mm}
  \label{tab:result-family}
\end{table*}

\subsection{Setup}
\label{ssec:setup}

\noindent
\textbf{Models and Datasets.}
We center our study on Qwen2.5~\cite{qwen2d5}, using Qwen2.5-3B-Instruct as the teacher model and Qwen2.5-0.5B and 1.5B as student models.
To cover other families, we also evaluate Qwen3 (4B$\rightarrow$0.6B) and Llama3.2 (3B$\rightarrow$1B).
All student models are initialized from base pretrained checkpoints, whereas the corresponding teacher models use instruction-tuned variants.
For the training dataset, we use SEA-Instruct, which covers seven SEA languages: Indonesian, Vietnamese, Thai, Tamil, Tagalog, Malay, and Burmese.
The dataset contains open-source prompts, each paired with a synthetic response and quality estimate.
We sample 4,000 high-quality examples per language, as labeled by the original dataset, yielding 28,000 samples.
Random sampling constraints preserve the distribution of domains, task types, and prompt complexity.

\noindent
\textbf{Framework Setup.}
The cross-lingual verbalizer in DuDi incorporates 7 SEA languages and English. 
%
%
During on-policy training, the target response language for the student is uniformly sampled from this set, excluding the original language of the training sample. 
In addition, we adopt a two-stage training framework in which the base model is first SFT on the SEA-Instruct dataset (cold-start SFT), after which DuDi training is initialized from the resulting SFT checkpoints.
For the sequence-level objective, the reference policy $\pi_\text{Ref}$ is set as the cold-start SFT checkpoint. 
The importance of cold-start SFT initialization is further discussed in Section~\ref{sec:cold_start}.
Additional details regarding the training configurations are provided in Appendix~\ref{appendix:config}.



\subsection{Evaluation}

We evaluate on SEA-HELM~\cite{sea-helm}, which covers multiple Southeast Asian languages and diverse tasks.
The languages are Indonesian, Vietnamese, Thai, Tamil, Tagalog, Malay, and Burmese.
SEA-HELM includes natural language understanding (NLU), natural language generation (NLG), natural language reasoning (NLR), safety, linguistic diagnostics, instruction following, and Southeast Asian knowledge.
All results are averaged over four seeds.
We also use Almost Stochastic Order (ASO)\footnote{\tiny Using the implementation from~\citet{ulmer2022deep}}~\cite{del2018optimal,dror2019deep} to test statistical significance between DuDi and the second-best performing framework.

\subsection{Competitive Methods} 
We compare DuDi with all comparative methods for SLMs, as we discussed in Section~\ref{subsec:background}.
For off-policy fine-tuning, we use SFT and DFT \citep{dft} as standard fine-tuning baselines. 
For iterative fine-tuning, we adopt SPIN \citep{spinwork} as a representative sequence-level optimization method. 
For self-distillation strategies, we include SDFT \citep{sdft}, which uses temporary prefix prompting for self-guided on-policy refinement.
For knowledge distillation, we evaluate SeqKD~\citep{seqkd} and GKD~\citep{gkd}.
Furthermore, 
all methods, except off-policy fine-tuning,
are initialized from a cold-start SFT checkpoint, following the same setup as DuDi. Additional implementation details for all competitive methods are provided in Appendix~\ref{appendix:config}.

\section{Main Results}
\label{sec:main_result}

\paragraph{DuDi outperforms all methods.}
Overall, DuDi achieves the strongest performance among all frameworks.
As shown in Table~\ref{tab:main-result}, DuDi obtains the highest average score of 10.1 across seven SEA languages, statistically significantly outperforming the strongest baseline, DFT, by 0.4 points and SPIN by 0.6 points.
DuDi ranks first in Vietnamese, Thai, and Malay, and second in Indonesian, Tamil, and Tagalog, demonstrating consistent improvements across both high- and mid-resource SEA languages.
The only exception is Burmese, where the gain is limited, likely due to the smaller teacher-student gap and teacher performance in this language.

\paragraph{Comparison with off-policy fine-tuning.}
DFT is a strong off-policy baseline, achieving the second-best overall score, only 0.4 points below DuDi.
It outperforms the larger model in Tamil and Burmese, suggesting that direct fine-tuning is beneficial when the teacher is unreliable. However, DFT does not consistently surpass SFT across languages and is limited as a cold-start initialization for further distillation (Reported in Appendix~\ref{sec:analysis-limit-dft}).


\noindent \textbf{Contrast with other methods.}
SPIN emerges as the third strongest approach, suggesting that self-play remains effective. GKD is also a strong KD baseline, achieving 9.0 on average, while SeqKD performs moderately but remains below both GKD and SPIN. In contrast, SDFT substantially underperforms despite its self-distillation design; its English-based verbalizer limits the teacher-student transferability. Further analyses regarding verbalizer configurations and teacher–student knowledge transferability are detailed in Sections \ref{sec:ablation-choice} and \ref{ssec:analysis-topk}.


\paragraph{Robustness across model families.}
DuDi also generalizes consistently across different teacher--student configurations.
As shown in Table~\ref{tab:result-family}, DuDi achieves the best average performance for all three student model families.
Compared with GKD, the strongest prior baseline, DuDi improves average performance by 5.2\%, 15.6\%, and 7.5\% on Qwen2.5-1.5B, Qwen3-0.6B-Base, and Llama3.2-1B, respectively.
These gains demonstrate that DuDi remains effective across variations in model scale, version, and architecture family.

\paragraph{Stability over competing methods.}
Among prior methods, GKD is the most competitive baseline and consistently outperforms SPIN and DFT across model families.
However, SPIN exhibits unstable behavior, it provides little improvement over SFT on Qwen2.5-1.5B and Llama3.2-1B, and only modest gains on Qwen3-0.6B-Base.
DFT is even less stable, failing to outperform SFT and collapsing on Llama3.2-1B.
In contrast, DuDi consistently improves over these baselines, highlighting its robustness over comparative fine-tuning approaches.

\section{Ablation Studies}
\label{sec:ablation}


\subsection{Critical Components}
\label{sec:ablation-components}

DuDi comprises sequence-level supervision, off-policy KD, and on-policy KD with a cross-lingual verbalizer.
We perform an ablation study by removing one component at a time and measuring its impact on overall performance.
Additionally, we examine whether excluding English from the set of target responses leads to performance degradation.

Table~\ref{tab:ablation-components} reports the ablation results.
Eliminating the sequence-level loss yields only a minor performance decline, though it still helps align the student with the ground-truth trajectory.
Excluding off-policy KD causes the largest performance drop, highlighting its critical role in providing ground-truth supervision and guiding SLMs toward the target trajectory.
Disabling on-policy KD results in the second-largest degradation, as the teacher can no longer refine student-generated responses.
%
%
%

For the verbalizer component,
their absence noticeably reduces performance by 5\%.
Lastly, removing English generation from the cross-lingual verbalizer degrades performance, suggesting the inclusion of the English language facilitates better cross-lingual transfer.
Overall, all components are complementary and jointly contribute to DuDi's optimal performance.

\begin{table}[htbp]

\vspace{-2mm}
  \centering
  \scalebox{0.65}{
  \begin{tabular}{l|rrr}
    \hline
    \textbf{Method} & \textbf{SEA-HELM} & \textbf{$\Delta$} & \textbf{\% Difference} \\
    \hline
    DuDi & 10.1  &  - &  - \\
    \hdashline
    w/o sequence & 9.7  & -0.4  & -4.7\% \\
    w/o off-policy KD &  7.6 & -2.5 & -24.6\% \\
    w/o on-policy KD &  9.5 & -0.6 & -6.6\% \\
    w/o verbalizer &  9.6 & -0.5 & -5.0\% \\ 
    w/o English & 9.8  & -0.3 & -3.3\% \\
    \hline        
  \end{tabular}
  }
  \vspace{-2mm}
  \caption{
  Ablation results of DuDi's components.
  }
  \label{tab:ablation-components}
  
\vspace{-5mm}
\end{table}

\subsection{Design Choices}
\label{sec:ablation-choice}

\noindent \textbf{Verbalizer Modes.} 
As described in Section~\ref{sec:DuDi}, DuDi uses a cross-lingual verbalizer.
We compare three variants: \textit{English}, following~\citet{sdft}, where verbalizer prompt template are English;
\textit{Multilingual}, where $l_z$ is the sample language $l$, and
\textit{Mix}, which uses 50\% multilingual mode for native-language learning and 50\% cross-lingual mode for knowledge transfer. Templates of individual verbalizers are provided in Appendix~\ref{appendix:verb}.

Table~\ref{tab:ablation-verbalizer-objectives} shows that our cross-lingual verbalizer is the only variant outperforming the no-verbalizer baseline.
The multilingual verbalizer performs worst overall, while the English-only verbalizer slightly surpasses it despite lacking language-specific supervision.
The mixed verbalizer further improves upon the multilingual setting, underscoring the importance of cross-lingual verbalization.
Overall, these results support the effectiveness of the proposed cross-lingual verbalizer in facilitating teacher-student knowledge transfer.


\noindent \textbf{KD Objective.} 
We compare DuDi’s reverse KL objective against forward KL and Jensen-Shannon Divergence (JSD), which interpolates between the two.
Table~\ref{tab:ablation-verbalizer-objectives} shows that reverse KL achieves the best SEA-HELM score (10.1), substantially outperforming JSD and forward KL, corresponding to relative drops of 23.8\% and 37.8\%, respectively.
The weaker performance of forward KL may stem from teacher-student mismatch and overestimation of low-probability regions in the teacher distribution \cite{gu2024minillm}. In contrast, reverse KL yields more stable and effective knowledge transfer.


\begin{table}[htbp]
  \centering
  \scalebox{0.65}{
  \begin{tabular}{l|rrr}
    \hline
    \textbf{Method} & \textbf{SEA-HELM} & \textbf{$\Delta$} & \textbf{\% Difference} \\
    \hline
    DuDi & 10.1  &  - &  - \\
    \hline\hline
    \multicolumn{4}{c}{\textit{Verbalizer Mode}} \\
    \hline
    No verbalizer &  9.6 & -0.5 & -5.0\% \\
    English & 8.2  & -1.9 & -18.8\% \\
    Multilingual & 7.9 & -2.2 & -21.8\% \\
    Mix & 9.0 & -1.1 & -10.9\% \\
    \hline\hline
    \multicolumn{4}{c}{\textit{KD Objective}} \\
    \hline
    Forward-KL &  6.3 & -3.8 & -37.8\% \\
    JSD & 7.7  & -2.4 & -23.8\% \\    
    \hline        
  \end{tabular}
  }
  \vspace{-2mm}
  \caption{
  Evaluation of performance under different verbalizers and knowledge distillation objectives.
  }
  \vspace{-4mm}
  \label{tab:ablation-verbalizer-objectives}
\end{table}


\section{Analyses}
\label{sec:analysis}

To further investigate the properties of DuDi, we conduct two analyses.
In Section~\ref{ssec:analysis-topk}, we examine how cross-lingual verbalizers facilitate the transferability between teacher and student.
Subsequently, in Section~\ref{sec:cold_start}, we analyze the role of cold-start fine-tuning and demonstrate its importance for effective multilingual knowledge distillation.

\subsection{Why DuDi Verbalizer is Optimal?}
\label{ssec:analysis-topk}

To better understand the effectiveness of the proposed cross-lingual verbalizer, we employ the overlap ratio analysis \cite{rethinking-opd}, which evaluates the top-k overlap between student and teacher output logits during on-policy token-level distillation. This metric quantifies the degree of agreement between the two distributions, thereby facilitating the on-policy gradient signal to the student model.
%
%

\begin{figure}[h!]
\centering
\includegraphics[width=0.9\columnwidth]{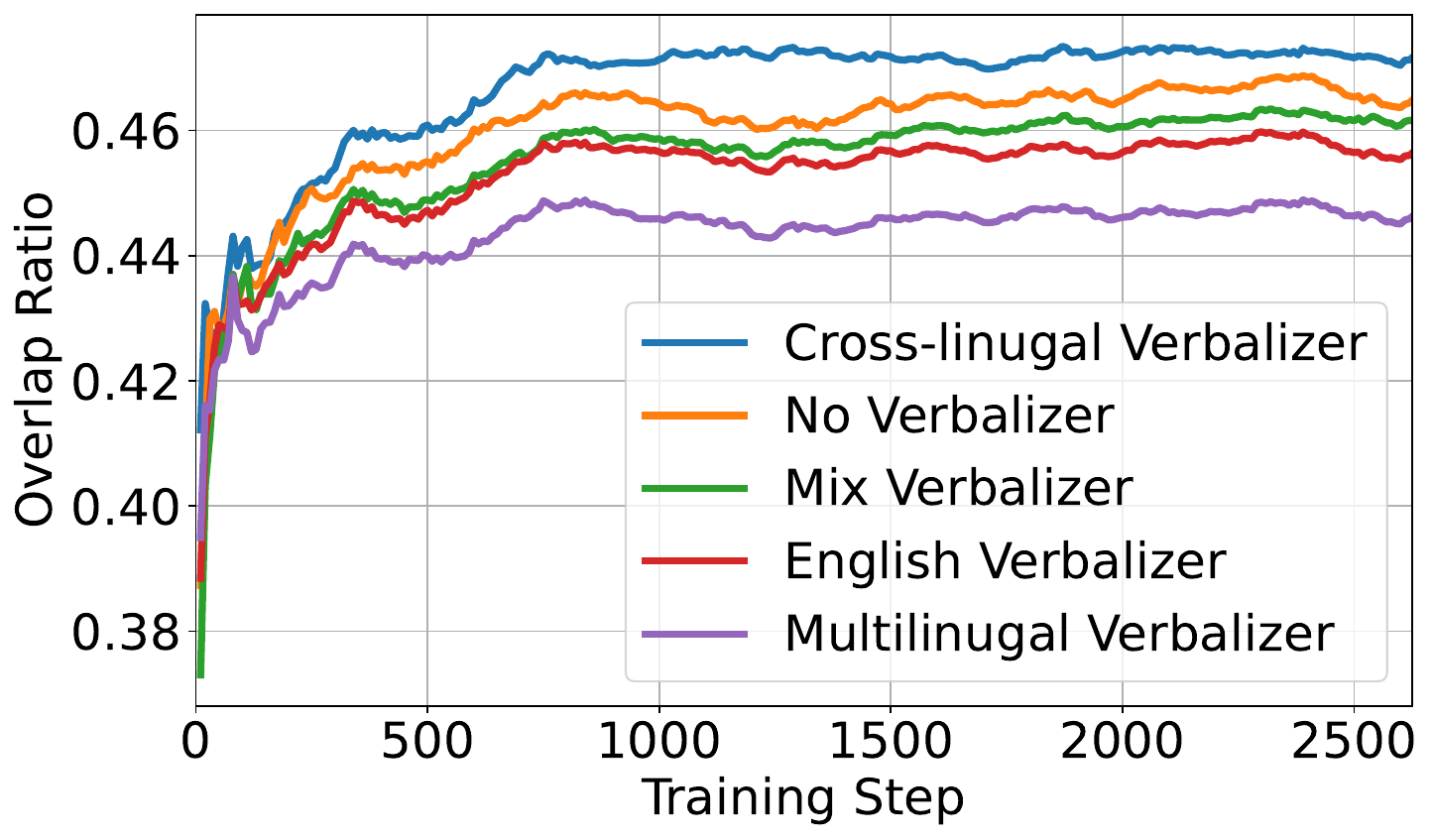}
\vspace{-3mm}
    \caption{Overlap ratio between teacher and student logits during on-policy rollouts across different verbalizers. The mix verbalizer denotes a uniform random combination of multilingual and cross-lingual verbalizers.}
    \vspace{-5mm}
    \label{fig:verbalizer_overlap_ratio}
\end{figure}

As shown in Figure~\ref{fig:verbalizer_overlap_ratio}, the proposed cross-lingual verbalizer achieves the highest overlap ratio throughout training, indicating that cross-lingual rollouts provide informative supervision signals for on-policy distillation. For the remaining variants, the overlap ratio ranking follows the performance trend in Table~\ref{tab:ablation-verbalizer-objectives}, with no verbalizer performing second best, followed by the mix, English, and multilingual verbalizers. Notably, the results suggest that multilingual verbalizers can increase teacher-student mismatches, highlighting the challenges of verbalizer design in multilingual settings.

%

\subsection{Why is Cold-Start SFT Important?}
\label{sec:cold_start}

DuDi is trained under a cold-start SFT setting, where models are initialized from SFT checkpoints using ground-truth responses as supervision. In parallel, \citet{rethinking-opd,sftonbasemodel} adopts a related initialization strategy based on teacher-generated responses rather than ground-truth data, demonstrating that SFT initialization improves the effectiveness of on-policy distillation.
As shown in Table~\ref{tab:cold-start}, training DuDi directly from the base model without cold-start initialization results in the weakest performance. Although initialization with teacher-generated outputs yields moderate improvements, it remains lower than initialization using the original ground-truth training data.
These findings indicate that cold-start initialization also plays an important role in effective multilingual knowledge distillation for SLMs.
Furthermore, similar trends are consistently observed across all comparative methods, as reported in Appendix~\ref{appendix:additional-results}.


\vspace{-2mm}
\begin{table}[h!]
  \centering
  \resizebox{\columnwidth}{!}{
  \begin{tabular}{l|rrr}
    \hline
    \textbf{Method} & \textbf{SEA-HELM} & \textbf{$\Delta$} & \textbf{\% Difference} \\
    \hline
    DuDi & 10.1  &  - &  - \\
    \hdashline
    Teacher generated & 9.1  & -1.0 & -10.2\% \\    
    No cold-start &  8.6 & -1.5 & -15.1\% \\
    \hline        
  \end{tabular}
  }
  \vspace{-2mm}
  \caption{
  Evaluation of downstream performance under different cold-start setting initialization.
  }
  \label{tab:cold-start}
\end{table}
\vspace{-5mm}

\section{Conclusion}
\vspace{-2mm}

In this work, we introduced DuDi, a general-purposed multilingual distillation framework for SLMs that integrates sequence-level and token-level supervisions, along with a cross-lingual verbalization mechanism.
Extensive experiments across diverse model families and parameter scales demonstrate that DuDi consistently achieves the highest average SEA-HELM performance, substantially outperforming strong fine-tuning and distillation baselines.
Ablation studies further indicate that jointly optimizing sequence-level and token-level objectives, together with the proposed verbalizer design, yields complementary benefits.
Our analysis also shows that the cross-lingual verbalizer improves teacher-student knowledge transferability.
Overall, DuDi offers an effective fine-tuning framework for SLMs in a multilingual environment.
In addition, for the open-research purpose, we will release all artifacts in this paper, including training code, datasets, and models.

\section*{Limitations}
The experimental setup of this study primarily focuses on Southeast Asian (SEA) languages, with models trained on SEA-Instruct and evaluated using SEA-HELM \cite{sea-helm}, which covers seven SEA languages across a diverse set of tasks. Consequently, the findings related to DuDi may not generalize to tasks beyond those included in the current evaluation framework. Nevertheless, SEA-HELM remains a gold-standard benchmark for the comprehensive evaluation of language model capabilities in SEA languages. Future work will focus on extending both the training data and evaluation benchmarks to encompass a wider range of contemporary language modeling tasks.

Another limitation of this work concerns the availability of suitable teacher models. In particular, the teacher model must be more capable, typically larger in scale than the student model while sharing the same output vocabulary space. This constraint arises because the divergence function employed in knowledge distillation requires aligned teacher–student logit dimensions.

\section*{Acknowledgments}
This project is supported by the National Research Foundation, Singapore under its National Large Language Models Funding Initiative. Any opinions, findings and conclusions or recommendations expressed in this material are those of the author(s) and do not reflect the views of National Research Foundation, Singapore. 
We thank Trevor Cohn for his helpful feedback, and Ngee Chia Tai and Raymond Ng for their support and valuable comments.
%





\bibliography{custom}

\appendix
\section{Related Work}
\label{sec:related-work}

\subsection{Knowledge Distillation}
Knowledge distillation (KD) \citep{hintonkd} is a training paradigm that transfers knowledge from a larger teacher model to a smaller student model, enabling compact models to benefit from the capabilities of stronger models. 
Early sequence-level distillation methods, including SeqKD \citep{seqkd} and ImitKD \citep{ImitKD}, demonstrate that teacher-generated outputs provide effective supervision signals for student training. 
Subsequent work has focused on improving the stability and efficiency of distillation. 
For example, MiniLLM \citep{gu2024minillm} introduces a policy-gradient-based framework that mitigates the high variance commonly encountered in reinforcement learning optimization. 
Similarly, \citet{xu2025speculative} combines static datasets with on-policy distillation through speculative decoding for synthetic data generation. 
Among recent approaches, DistiLLM \citep{DistiLLM-2} achieves strong performance and training efficiency by employing symmetric KL divergence together with an adaptive off-policy distillation strategy. 

More recent studies extend distillation to on-policy settings, where student models learn directly from their own generated responses. 
For instance, \citet{gkd} proposes on-policy optimization objectives based on reverse KL divergence and Jensen-Shannon divergence (JSD). 
Beyond conventional teacher distillation paradigms, self-distillation methods exploit model-generated responses as supervision signals to iteratively improve reasoning and downstream capabilities. 
Specifically, offline self-distillation methods \citep{sdft-acl} utilize self-generated responses from a seed model to better align the model with its own output distribution, whereas online variants \citep{sdft} employ in-context learning to acquire new capabilities while retaining the original competencies of the base model, alongside related reinforcement learning formulations proposed by \citet{sdpo}.
In parallel, sequence-level optimization methods such as self-play approaches like SPIN \citep{spinwork}, and DistiLLM-2 \citep{DistiLLM-2} leverage contrastive objectives and trajectory-level regularization to enhance sample efficiency in reasoning-focused tasks.


\subsection{Multilingual Distillation}
Recent studies have primarily focused on constructing high-quality datasets for training smaller models on reasoning tasks. 
\citet{d-cot-nr} propose a distillation framework that transfers teacher capabilities through response generation, leveraging both positive and negative rationales to fine-tune smaller models for question-answering tasks. 
From a data-centric perspective, \cite{SDRRL-multi-self-distill-resource} present self-distillation from transferring rich-resources to low-resources and MathOctopus \citep{mathoctopus} translates mathematical training data into target languages to improve multilingual mathematical reasoning performance. 
On the architectural side, \citet{langbridge} introduce a multilingual encoder integrated with reasoning-capable LLMs for solving multilingual mathematics problems. 
In addition, \citet{mwork} investigate the disentanglement of language and reasoning by identifying and exploiting language-specific neurons, thereby enhancing multilingual capabilities. 
Similarly, \citet{lesslanguageismore} propose causal intervention methods to improve downstream multilingual reasoning performance. 
Despite these advances, existing knowledge distillation frameworks remain largely centered on English settings and single task-specific setting, while general-purpose training frameworks for multilingual LLMs are still underexplored.

\section{Training Configuration}
\label{appendix:config}

We fine-tune all models and all methods with a learning rate of 2e-5, batch size of 32, and 3 training epochs. The maximum sequence length is set to 1024 tokens, while on-policy rollouts are limited to a maximum of 256 generated tokens. 
For off-policy fine-tuning, we apply loss only on assistant tokens to better align with the next-token prediction objective of language models. 
In the on-policy rollout, following \citet{gkd}, we enable stochastic exploration by setting\footnote{\tiny\url{https://huggingface.co/docs/transformers/main_classes/text_generation}} \texttt{do\_sample=True} and \texttt{top-k=0}, with temperature 0.9 for all approaches.
All knowledge distillation objectives use a temperature of 1.0.
For hyperparameters in DuDi, we use fixed values of $\lambda=0.5$ and $\alpha=0.1$ across all experiments without per-model tuning.

For other methods, we evaluate SeqKD \citep{seqkd} with $\lambda=1.0$, corresponding to training solely on teacher-generated responses, and GKD \citep{gkd} using its default setting of $\lambda=0.5$. Both SeqKD and GKD use JSD as the objective.
We implement DuDi and all comparative methods using the TRL trainer framework \cite{vonwerra2020trl}, except for SDFT \cite{sdft}, where we adopt the original codebase provided by the authors.

\section{Computing Resources}
\label{appendix:computing-resources}

We trained small models on 8× NVIDIA H200 (140GB) GPUs, completing fine-tuning within approximately 7 hours for SDFT, SeqKD, SPIN, GKD, and DuDi, and within an hour for SFT and DFT. For evaluation, we used 2× NVIDIA H200 (140GB) GPUs, completing SEA-HELM within 1 hour. In total, our experiments required approximately 1,704 GPU hours.

\section{Top-K Overlap Analysis Details}
\label{appendix:topk-detail}

For visualization purposes, the top-k overlap analysis presented in this paper is smoothed using an exponential moving average with a smoothing factor of $0.99$.\footnote{\tiny \url{https://docs.wandb.ai/models/app/features/panels/line-plot/smoothing.}}

\section{Verbalizer Template}
\label{appendix:verb}
\begin{figure}[h!]
\centering
    \includegraphics[width=\columnwidth]{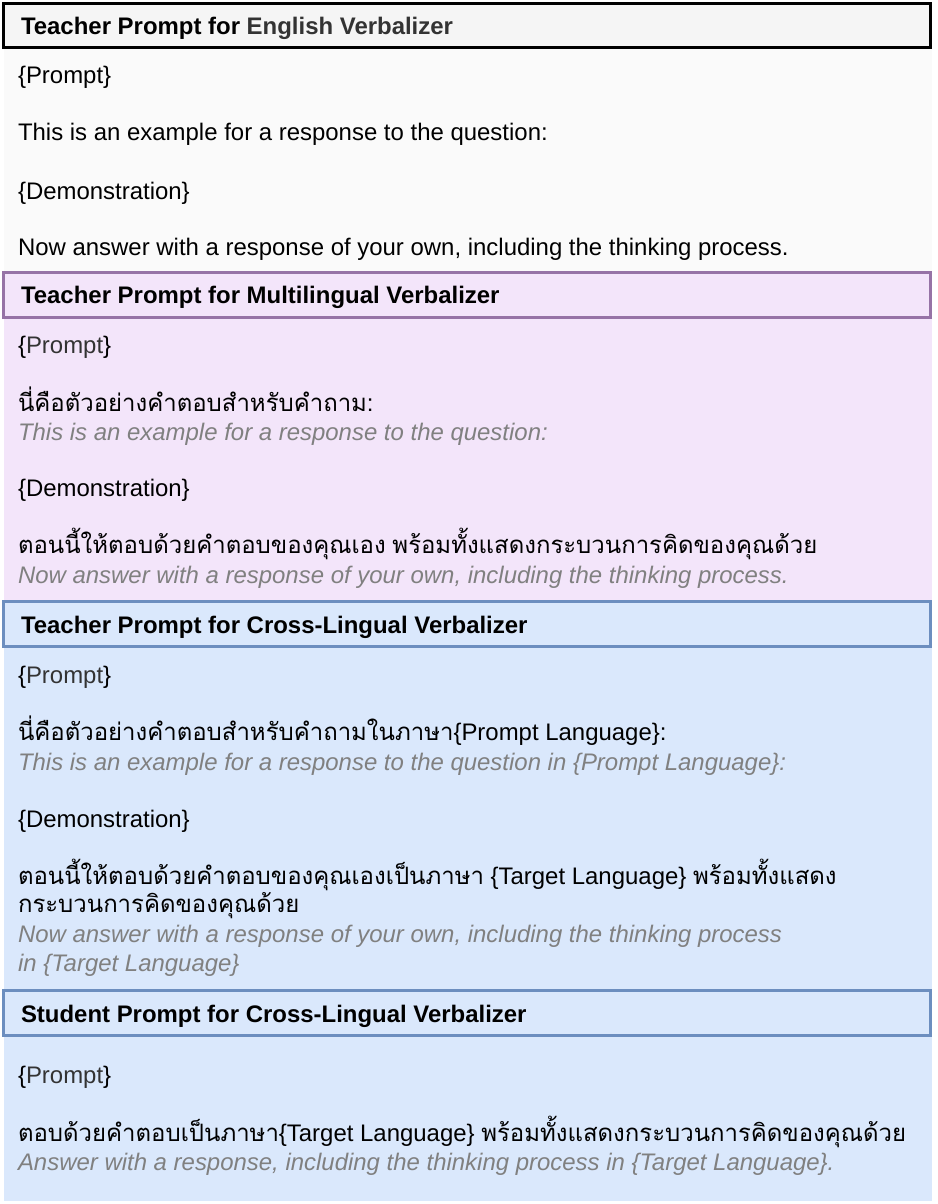}
    \caption{Comparison of verbalizer templates for teacher prompt, including the English verbalizer from \cite{sdft}, our extended multilingual verbalizer, and the proposed cross-lingual verbalizer with its corresponding student prompt example in Thai.}
    \label{fig:verbalizer_template}   
    
    \vspace{-2mm}
\end{figure}

For each training sample $(x, y, l)$, there are three verbalizer modes: \textbf{(i) English}, where verbalizer prompt template are English, following~\citet{sdft}; \textbf{(ii) Multilingual}, where $l_z = l$; and \textbf{(iii) Cross-lingual}, where verbalizer prompt template language matches the sample’s native language $l$, while $l_z$ is sampled uniformly from the set consisting of English and all training languages excluding $l$. Templates for all verbalizer mode are shown in Figure~\ref{fig:verbalizer_template}.

\section{Limitations of DFT as Cold-Start}
\label{sec:analysis-limit-dft}

\begin{table}[h!]
  \centering
  \resizebox{\columnwidth}{!}{
  \begin{tabular}{l|rrrrrrr|l}
    \hline
    \textbf{Method} & \textbf{id} & \textbf{vi} & \textbf{th} & \textbf{ta} & \textbf{tl} & \textbf{ms} & \textbf{my} & \textbf{Avg. ($\Delta$)} \\
    \hline
    SFT &  10.9 & 12.1  & 11.2  & 5.1  & 5.8  & 12.1  & 4.3  & 8.8 \\
    DFT & 9.3  & 11.6  & 10.2  & 10.0  & 6.1  & 12.4  & 8.3  &  9.7 \\
    \hline       
    SFT $\rightarrow$ SPIN & 12.6  & 11.4  & 10.0  & 5.3  & 8.0  & 14.7  & 4.8  & 9.5 \textcolor{AcademicGreen}{(+0.7)} \\
    DFT $\rightarrow$ SPIN & 11.5  & 8.5  & 9.1  & 9.3  & 5.0  & 12.4  & 8.2  & 9.1 \textcolor{AcademicRed}{(-0.6)} \\
    \hdashline    
    SFT $\rightarrow$ GKD & 11.7 & 13.2 & 10.9  & 4.9  & 4.8  & 13.6  & 3.7  & 9.0 \textcolor{AcademicGreen}{(+0.2)} \\
    DFT $\rightarrow$ GKD & 6.6  & 7.8  & 7.9  & 3.9  & 4.5  & 7.2  & 3.6  &  5.9 \textcolor{AcademicRed}{(-3.8)} \\
    \hdashline    
    SFT $\rightarrow$ DuDi & 12.8  & 14.6  & 11.5  & 5.4  & 6.4  & 14.6  & 3.4  & 10.1 \textcolor{AcademicGreen}{(+1.3)} \\
    DFT $\rightarrow$ DuDi & 7.9  & 10.8  & 8.9  & 6.4  & 4.9  & 9.1  & 4.1  & 7.4 \textcolor{AcademicRed}{(-2.3)} \\
    \hline
  \end{tabular}
  }
  \vspace{-2mm}
  \caption{
  Comparison of results using alternative SFT- or DFT-based checkpoints as a cold-start across different training frameworks.
  }
  \vspace{-2mm}
  \label{tab:analysis-dft}
\end{table}

As shown in Table~\ref{tab:main-result}, DFT outperforms SFT on downstream tasks.
Motivated by this, we examine whether DFT provides a better initialization checkpoint than SFT as a cold-start.
We compare off-policy fine-tuning (cold-start) initialized from SFT and DFT checkpoints across three methods: SPIN, GKD, and DuDi.
As shown in Table~\ref{tab:analysis-dft}, using a DFT checkpoint as the cold-start degrades performance across all three frameworks.
Thus, while DFT is a strong standalone baseline, it is less compatible with subsequent training and yields limited additive gains.

%

\begin{figure}[h!]
\vspace{-2mm}
\centering
\includegraphics[width=0.9\columnwidth]{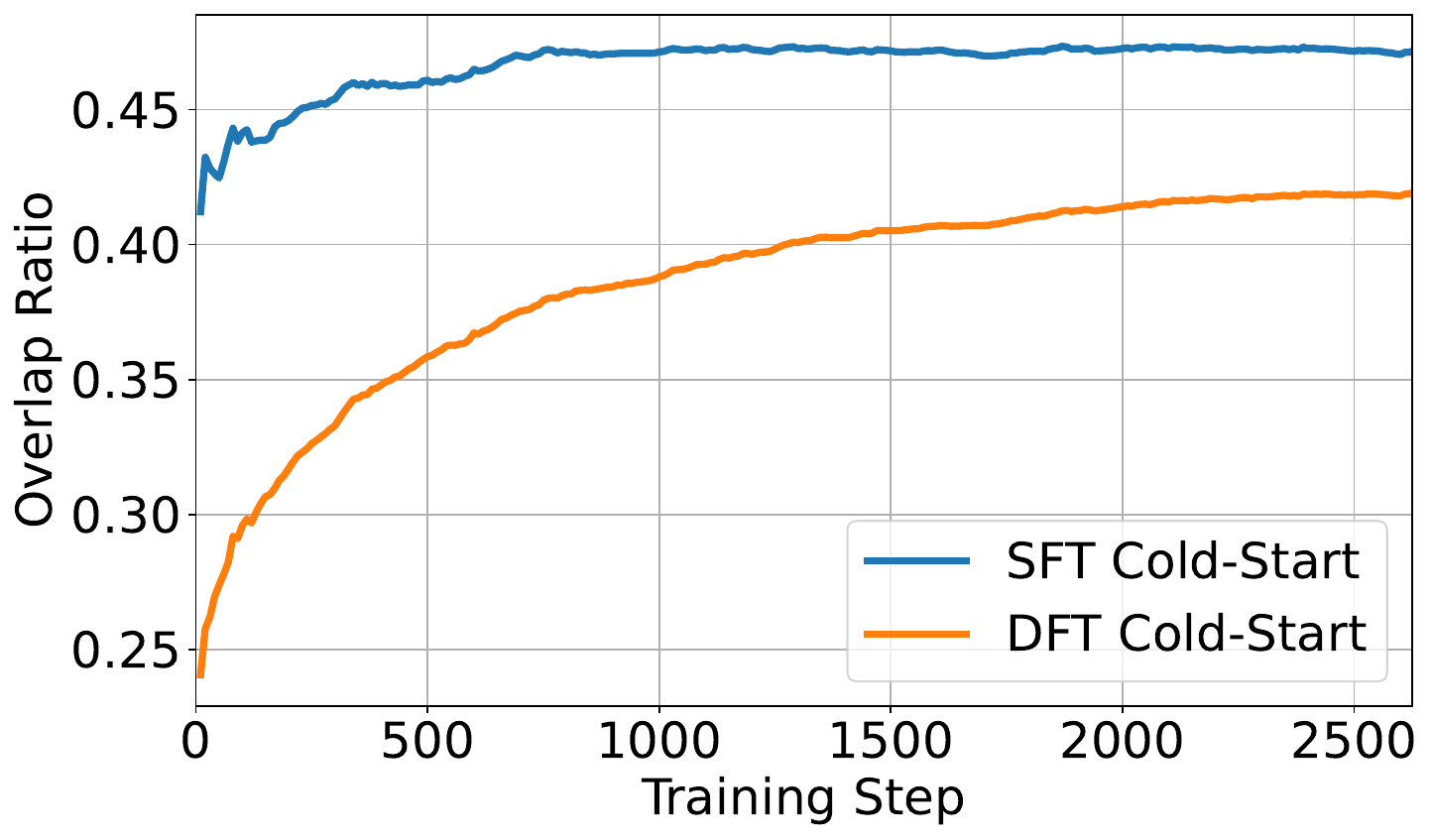}
\vspace{-4mm}
    \caption{
    Overlap ratio between student and teacher model logits for on-policy token-level distillation of SFT vs DFT as a cold-start.
    }
    \vspace{-4mm}
    \label{fig:dft_overlap_ratio}
\end{figure}

To better understand the underlying mechanism between SFT and DFT as the cold-start, we compare the top-k overlap ratio of SFT and DFT as the cold-start initializations for DuDi. As illustrated in Figure~\ref{fig:dft_overlap_ratio}, the cold-start DFT consistently exhibits a substantially lower overlap ratio than SFT throughout training across methods. According to \citet{rethinking-opd}, a low overlap ratio is indicative of degraded performance, suggesting that the cold-start DFT student assigns probability mass to a token set that is largely disjoint from that of the teacher. These findings indicate that DFT alters the token distribution too aggressively, biasing the model toward different token preferences.

\section{Additional Results}
\label{appendix:additional-results}

\noindent \textbf{Task-Level Performance.}
Table~\ref{tab:task-qwen3} shows downstream evaluation for Qwen2.5-0.5B, Qwen2.5-1.5B, Qwen3-0.6B-Base, and Llama-3.2-1 on multiple tasks in SEA-HELM, showing capability of each training methods. DuDi consistency achieves either the best or second-best performance across most tasks and model families.
Notably, several tasks, Natural Language Reasoning (NLR), Safety, Linguistic Diagnostics (LD), and Knowledge, exhibit near-zero to very low absolute performance for Qwen2.5-0.5B and Llama-3.2-1B (often below 3 points), despite the teacher demonstrating some capability on these tasks. In contrast, larger or more capable base models such as Qwen2.5-1.5B and Qwen3-0.6B-Base retain non-trivial performance on these tasks.
We hypothesize that this gap arises from limited coverage of these task distributions in the training data, resulting in weak teacher--student transfer and indicating that such tasks may require stronger base model capabilities rather than distillation alone.

\noindent \textbf{Cold-Start SFT.} Consistent with our findings in Section~\ref{sec:cold_start} and prior studies \citep{rethinking-opd,sftonbasemodel}, Table~\ref{tab:result-wo-sft} highlights the importance of cold-start SFT in knowledge distillation. The performance of both self-distillation methods, such as SDFT, and teacher-distillation approaches (SeqKD and GKD) declines when the student model is not initialized with cold-start SFT.

\begin{table*}[h!]
  \centering
  \scalebox{0.8}{
  \begin{tabular}{l|rrrrrrr|r}
    \hline
    \textbf{Method} & \textbf{NLU} & \textbf{NLG} & \textbf{NLR} & \textbf{Safety} & \textbf{LD} & \textbf{IF} & \textbf{Knowledge} & \textbf{Average}\\
\hline
Qwen2.5-3B-Instruct (Teacher) & 32.5 & 34.8 & 16.4 & 19.8 & 13.5 & 47.2 & 15.1 & 27.7 \\
\hline
\multicolumn{1}{l|}{Qwen2.5-0.5B (Student)} & \multicolumn{7}{c|}{} & \multicolumn{1}{c}{}\\
SFT & 7.2 & 12.5 & 0.4 & 0.2 & 0.0 & 28.8 & \underline{2.2} & 8.6 \\
DFT & 7.3 & 9.5 & 0.2 & 0.0 & \underline{0.4} & \textbf{38.5} & 0.9 & \underline{9.7} \\
\hdashline
SPIN & 7.5 & 13.9 & \textbf{1.1} & \textbf{0.4} & \textbf{0.5} & 28.6 & \textbf{5.2} & 9.5 \\    
\hdashline
SDFT & 1.1 & 3.8 & 0.0 & 0.0 & 0.0 & 16.6 & 0.4 & 3.6 \\
SeqKD & 3.2 & 9.5 & 0.0 & 0.0 & 0.0 & 22.4 & 1.3 & 6.1 \\
GKD & \underline{8.1} & \underline{15.6} & 0.3 & \underline{0.3} & 0.0 & 28.1 & 1.4 & 9.0 \\
DuDi (ours) & \textbf{9.0} & \textbf{15.8} & \underline{0.6} & 0.1 & 0.2 & \underline{33.6} & 1.4 & \textbf{10.1} \\
        \hline \hline
Qwen2.5-3B-Instruct (Teacher) & 32.5 & 34.8 & 16.4 & 19.8 & 13.5 & 47.2 & 15.1 & 27.7 \\
    \hdashline
    \multicolumn{1}{l|}{Qwen2.5-1.5B (Student)} & \multicolumn{7}{c|}{} & \multicolumn{1}{c}{}\\

    SFT & 22.0 & \textbf{26.3} & 6.2 & 3.6 & 0.6 & \underline{40.7} & 13.3 & 18.6 \\
    DFT & 17.4 & 9.9 & \textbf{11.9} & 4.5 & \textbf{4.1} & 35.2 & \textbf{15.4} & 15.6 \\
    SPIN & 20.4 & 23.7 & \underline{8.0} & 5.1 & 0.4 & 38.8 & \underline{14.6} & 18.3 \\
    GKD & \underline{24.9} & 22.0 & 7.8 & \underline{7.9} & 2.3 & 41.3 & 10.5 & \underline{19.1} \\
        DuDi & \textbf{26.5} & \underline{23.7} & 6.1 & \textbf{8.8} & \underline{2.5} & \textbf{44.1} & 10.2 & \textbf{20.1} \\
    
    \hline \hline
    Qwen3-4B (Teacher) & 58.6 & 25.2 & 47.3 & 37.9 & 36.2 & 68.3 & 33.1 & 45.6 \\
    \hdashline
    \multicolumn{1}{l|}{Qwen3-0.6B-Base (Student)} & \multicolumn{7}{c|}{} & \multicolumn{1}{c}{}\\
    SFT & 16.6  & 22.8 & 4.5 & 0.7 & 0.0 & 35.6 & 2.2 & 13.9 \\
    DFT & 20.7   & 8.0  & 7.6 & 2.0 & \underline{1.2} & 40.6 & \textbf{12.6} & 15.6 \\    
    SPIN & 20.0  & 18.6 & 5.6 & 0.9 & 0.2 & 36.3 & 2.3 & 14.3 \\
    GKD & \underline{24.6}  & \underline{26.2} & \underline{8.1} & \underline{3.0} & 0.0 & \underline{41.5} & 1.9 & \underline{18.0} \\
    DuDi & \textbf{28.5}  & \textbf{26.3} & \textbf{14.0} & \textbf{3.7} & \textbf{2.8} & \textbf{44.7} & \underline{5.5} & \textbf{20.8} \\
    \hline \hline
    Llama-3.2-3B-Instruct (Teacher) & 34.2 & 44.2 & 7.9 & 15.8 & 1.0 & 47.2 & 16.0 & 27.7 \\
    \hline
    \multicolumn{1}{l|}{Llama-3.2-1B (Student)} & \multicolumn{7}{c|}{} & \multicolumn{1}{c}{}\\
    SFT & 1.5 & 6.2 & 0.0 & 0.0 & 0.0 & 20.0 & 0.3 & 4.7 \\
    DFT & 0.0 & 3.3 & 0.0 & 0.0 & 0.0 & 0.0 & 0.0 & 0.6 \\
    SPIN & 0.0 & 21.7 & 0.0 & 0.0 & 0.0 & 0.0 & 0.0 & 3.7 \\
    GKD & \underline{9.3} & \underline{23.1} & \underline{0.4} & \underline{0.5} & 0.0 & \underline{28.1} & \textbf{2.2} & \underline{10.6} \\
    DuDi & \textbf{9.7} & \textbf{24.0} & \textbf{0.4} & \textbf{2.9} & \textbf{0.4} & \textbf{30.0} & \underline{1.2} & \textbf{11.4} \\
    \hline  

  \end{tabular}
  }
  \vspace{-3mm}
  \caption{
  Task evaluations for Qwen2.5-0.5B, Qwen2.5-1.5B, Qwen3-0.6B-Base, and Llama-3.2-1B.
  }
  \vspace{-5mm}
  \label{tab:task-qwen3}
\end{table*}

\begin{table*}[h!]
  \centering
  \scalebox{0.8}{
  \begin{tabular}{l|rrrrrrr|r}
    \hline
    \textbf{Method} & \textbf{Indonesian} & \textbf{Vietnamese} & \textbf{Thai} & \textbf{Tamil} & \textbf{Tagalog} & \textbf{Malay} & \textbf{Burmese} & \textbf{Average}\\
    \hline
    SDFT & 4.2 & 4.6  & 4.7  & 2.3  & 3.3  & 5.1  & 1.3  & 3.6  \\    
    SDFT w/o SFT & 0.7  & 0.7  & 1.8  &  1.1 & 0.6  & 0.9  & 1.1  & 1.0  \\
    \hdashline
    SeqKD & 7.1 & 9.1  & 7.7  & 4.2  & 3.4  & 8.1  & 2.9  &  6.1 \\    
    SeqKD w/o SFT & 4.4  & 7.2  & 6.1  & 3.6  & 2.1  & 6.1  & 3.2  & 4.7  \\
    \hdashline
    GKD & 11.7  & 13.2  & 10.9  & 4.9  & 4.8  & 13.6  & 3.7  & 9.0  \\
    GKD w/o SFT & 6.5  & 7.5  & 7.2  & 3.9  & 3.5  & 7.7  & 3.3  & 5.6  \\
    \hdashline
    DuDi & 11.7 & 14.4 & 12.8  & 6.5  & 6.1  & 14.8 & 4.6  & 10.1 \\
    DuDi w/o SFT & 9.4 & 13.7 & 9.7 & 7.2 & 5.3 & 10.8 & 4.1 & 8.6 \\    
    \hline
  \end{tabular}
  }
  \vspace{-3mm}
  \caption{
  Results of comparative knowledge distillation methods without cold-start SFT.
  }
  \vspace{-3mm}
  \label{tab:result-wo-sft}
\end{table*}

\end{document}